\documentclass[letterpaper, 10 pt, conference]{ieeeconf}

\usepackage{color, soul}
\usepackage{cite}
\usepackage{amsmath, amssymb, amsfonts}
\usepackage{algorithmic}
\usepackage{graphicx}
\usepackage{textcomp}
\usepackage{comment}
\usepackage{tabu}
\usepackage{booktabs}
\usepackage{multirow}
\usepackage{multicol}
\usepackage{url}
\usepackage{array}   
\usepackage{fancyhdr}

\IEEEoverridecommandlockouts 

\usepackage{graphics} 
\usepackage{lipsum}  

\title{\LARGE \bf Close-Fitting Dressing Assistance Based on State Estimation of
\\Feet and Garments with Semantic-based Visual Attention
}

 \author{Takuma Tsukakoshi*, Tamon Miyake*, Tetsuya Ogata, Yushi Wang, Takumi Akaishi, and Shigeki Sugano
 \thanks{This work was supported by JST Moonshot R and D, Grant No. JPMJMS2031.}
 \thanks{$*$First two authors contributed equally. }
 \thanks{Takuma Tsukakoshi (Corresponding author) and Takumi Akaishi are with Creative Science and Engineering Faculty, Waseda Univ., Tokyo, Japan.
Tamon Miyake and Yushi Wang are with Future Robotics Organization, Waseda Univ., Tokyo, Japan. Tetsuya Ogata and Shigeki Sugano are with Faculty of Science and Engineering, Waseda
 Univ., Tokyo, Japan. Tetsuya Ogata is also with The
National Institute of Advanced Science and Technology, Tokyo, Japan.}
} 

\fancypagestyle{firstpage}{
  \fancyfoot[C]{\footnotesize IEEE Robotics and Automation Letters (RA-L)}
}

\begin{document}
    \maketitle
    \thispagestyle{empty}
    \pagestyle{empty}
    \begin{abstract}
        As the population continues to age, a shortage of caregivers is expected in the future.
        Dressing assistance, in particular, is crucial for opportunities for social participation. Especially dressing close-fitting garments, such as socks, remains challenging due to the need for fine force adjustments to handle the friction or snagging against the skin, while considering the shape and position of the garment. This study introduces a method that uses multi-modal information including not only the robot's camera images, joint angles, joint torques, but also tactile forces for proper force interaction that can adapt to individual differences in humans. Furthermore, by introducing semantic information based on object concepts, rather than relying solely on RGB data, it can be generalized to unseen feet and background. 
        In addition, incorporating depth data helps infer relative spatial relationship between the sock and the foot.
        To validate its capability for semantic object conceptualization and to ensure safety, training data were collected using a mannequin, and subsequent experiments were conducted with human subjects. In experiments, the robot successfully adapted to previously unseen human feet and was able to put socks on 10 participants, achieving a higher success rate than Action Chunking with Transformer and Diffusion Policy. These results demonstrate that the proposed model can estimate the state of both the garment and the foot, enabling precise dressing assistance for close-fitting garments.
    \end{abstract}

    \section{INTRODUCTION}
    As the population ages, the shortage of care workers is anticipated to worsen worldwide. One solution to this problem is developing autonomous caregiving robots. In physical caregiving, assistance with activities of daily living (ADL) is essential for a minimum standard of living.
    Autonomous dressing assistance is crucial to help care recipients participate in society because dressing is one of the most skill-intensive tasks for caregivers and one of the most frequent activities for social participation among ADLs.

    In garment manipulation, representing the state of the garment and the human body, as well as modeling their dynamics, is challenging due to occlusions and garment deformation. Moreover, the safety of humans should be considered. 
    Recent studies have addressed these challenges by leveraging multimodal information with deep learning techniques \cite{kotsovolis2024garment,kotsovolis2024model,sun2024force,wang2023one}.
    Previous research on dressing assistance has focused on garments, such as shirts and trousers, that have wide openings and allow for considerable misalignment between the garment and the body \cite{zhu2024you}.
    It is still difficult for a robot to make fine adjustments to the force applied when dressing, while accounting for the garment's shape and its friction/hooking with the skin. 
    Close-fitting dressing assistance, such as putting on socks, has not been fully realized.
    Furthermore, individual physique variations are also one of the factors that complicate dressing assistance, highlighting the need for the generalizability of robotic manipulation.

    \thispagestyle{firstpage}

    \begin{figure}
        \centering
        \vspace*{0.1cm}
        \includegraphics[width=5cm]{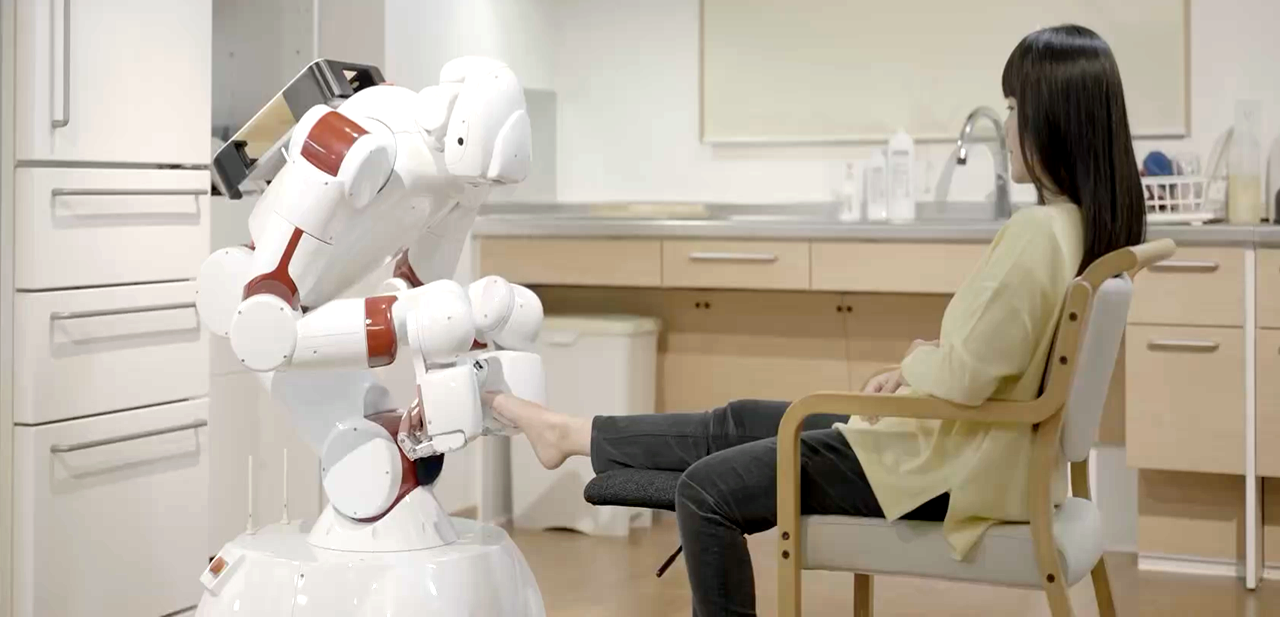}
        \vspace*{-0.18cm}
        \caption{The proposed model can learn object concepts, and the model trained on a mannequin's foot can be adapted to an untrained human foot.}
        \label{fig:outline}
       \vspace*{-0.66cm}
    \end{figure}

    In this study, we focus on sock-dressing assistance as a close-fitting dressing task.
    The sock-dressing assistance is important to help older people and people with physical limitations because they often have difficulty reaching down toward the lower parts of their body. 
    The objective is to generate a dressing motion based on the estimation of the state of the feet and socks, which is robust to individual differences (e.g., foot size, shape, skin tone, and flexibility) and generalizes to unseen backgrounds beyond the target objects. To achieve this, we propose a method that leverages semantic understanding of objects rather than relying solely on RGB images.
    Moreover, it is difficult to accurately reproduce the friction between human skin and socks in simulation, we aim to apply demonstration-based learning in the real world to sock-dressing assistance.

    In the process of dressing socks, simply pulling on the garment is insufficient. It is necessary to apply an appropriate pulling force—both in direction and magnitude—across the entire contact surface between the sock and the foot. Due to the elastic nature of socks, improper force direction can cause local slack in the material, preventing it from conforming smoothly to the foot's shape and often resulting in excess garment around the toe area. Conversely, if the sock is not adequately stretched before dressing, frictional resistance increases, leading to snagging and difficulty in progressing the dressing motion. Therefore, successful sock-dressing requires a careful balance between garment elasticity, friction at the sock-foot interface, and the directional control of the pulling force.
    To address these issues, we propose a new multimodal method based on joint states, finger pressure, and attention points with 3D features by combining a semantic mask image with a depth estimation. 
    The experiments show that the proposed model integrates a semantic-level understanding of the operating environment that is independent of the RGB values, and multimodal learning to accommodate individual differences, leading to a system with greater versatility and higher performance. 
    The contributions of this paper can be summarized as follows:  
    \begin{itemize}
    \item  Establishing an imitation learning method for policy of 
        close-fitting garment manipulation with proper force interaction adapting to individual differences in humans
        \item Evaluating generalizability of the proposed method, which was trained on a mannequin's foot for safety and applied to untrained real human feet.
        \item Evaluating robustness of the proposed method to color difference in background
    \end{itemize}

    \section{Related work}
    \subsection{Dressing assistance manipulation}

Numerous studies have explored robot-assisted dressing methods. 
Some studies assumed the human who has physical limitations but can move during the task. These studies typically tackle the challenge of adapting to user motion despite garment-induced occlusions, often integrating personalization techniques that consider individual preferences and physical limitations \cite{jevtic2019personalized,kapusta2019personalized,yamasaki2023realizing,canal2019adapting}.
Other types of study concentrated on garment manipulation prior to the dressing process, with a primary focus on grasping and unfolding garments to prepare them for a suitable dressing configuration \cite{zhang2020learning, zhang2023visual,zhang2022learning}.
During dressing, data-driven haptic perception could infer interaction between the garment and the human body \cite{kapusta2016data}.

Another line of research focuses on garment manipulation during the dressing process, proposing control strategies to guide the garments into a desired configuration around the human body.
Demonstration-based learning was performed for arm-dressing using dynamic movement primitives and body-dressing using a Bayesian method \cite{joshi2019framework}.
Model predictive control was applied based on point cloud observation of garments and body parts for garment opening insertion using graph convolutional network \cite{kotsovolis2024model} and using diffusion policy \cite{kotsovolis2024garment}.
A visual policy and a force dynamics model were combined to construct a motion policy for safe dressing assistance \cite{sun2024force}.
One policy for dressing different garments on people with diverse poses from partial point cloud observations was developed by leveraging policy distillation to combine policies of different pose sub-ranges \cite{wang2023one}.

     \begin{figure*}[h]
        \centering
        \includegraphics[width=\textwidth]{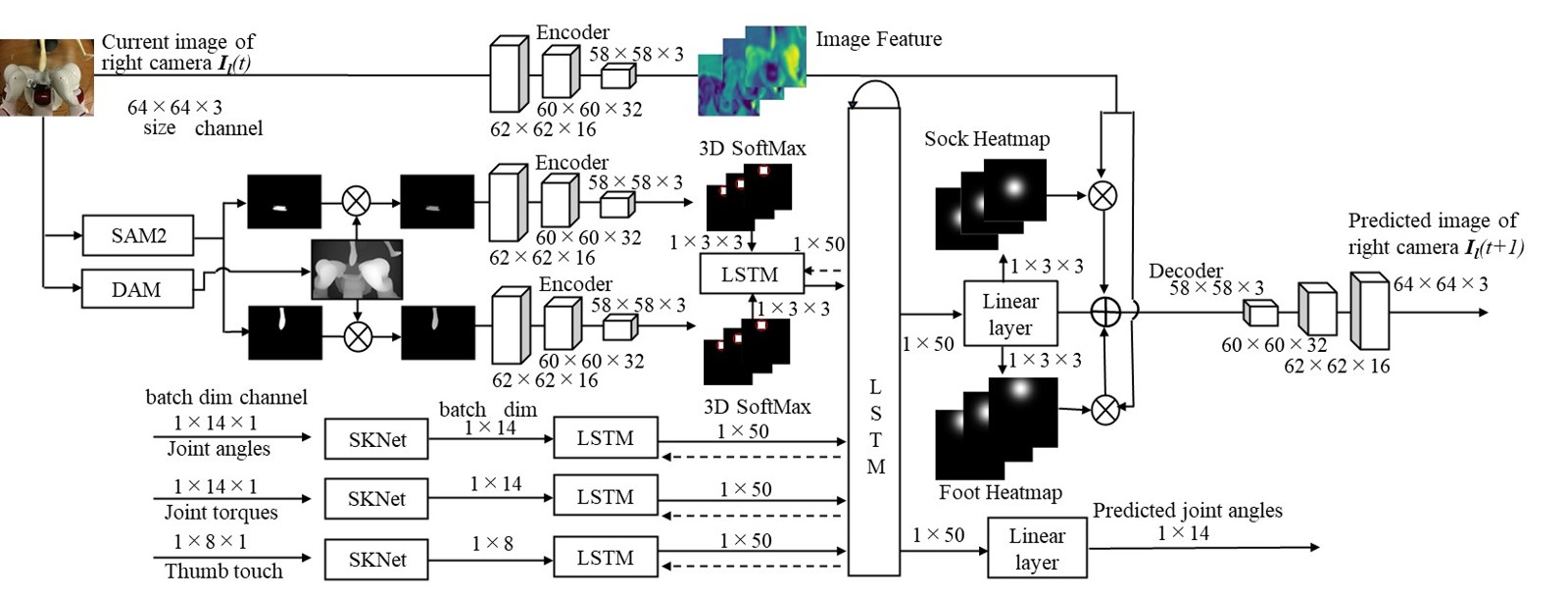} 
               \vspace*{-0.55cm}

        \caption{An overview of our proposed model.  The motion generation is based on the deep predictive learning model EIPL with hierarchical LSTM \cite{suzuki2023deep}. 
    The semantic extraction function obtains semantic information from images and estimates the 3D information of target objects. The features extracted from the semantic mask and depth estimation through the CNN are input into an attention mechanism called Spatial Softmax to obtain attention point information on the images. The visual attention points, joint angles, torques, and tactile information are input into a hierarchical LSTM, and the image and joint angle one step ahead are output.} 
        \label{fig:wide-image}
       \vspace*{-0.55cm}
    \end{figure*}

Previous studies mainly have addressed garments such as loose-fitting shirts and trousers.
Despite the robot's ability to make fine force adjustments while considering the shape of the garment and friction or entanglement with the skin, dressing assistance involving delicate, close-fitting garments, such as putting on socks, has yet to be realized.
It is necessary to account for individual differences in body part size, shape, color, and flexibility.

    \subsection{Imitation learning for deformable object manipulation}
Recently, imitation learning from demonstration is being applied to garment manipulation.
Garment manipulation is challenging due to the complicated dynamics of deformation, high-dimensional state representation, and perception complexities. Multimodal-based representation for manipulation policy, such as folding, unfolding, and smoothing, has been achieved \cite{seita2020deep, zhao2023learning,yang2016repeatable}.
Large language or vision language model facilitates more exhaustive tasks with various trajectory generation \cite{chen2025metafold,black2024pi_0}.
Many of these garment manipulation tasks can be addressed using operations based on simple pick-and-place. 
Although the imitation learning-based manipulation could maintain appropriate contact forces with objects \cite{adachi2018imitation,saito2021utilization},
It is still challenging to achieve force interaction for careful balance between fabric elasticity, friction at the sock-foot interface, and the directional control of the pulling force.

\section{Problem formulation and assumptions}
    Close-fitting garments, such as socks, remain a significant challenge due to their complex deformation and close contact with human skin.
    During dressing assistance, the garment often gets caught on the body, making it challenging to dress the human.
    Accurate force interaction, which guides the garment along the body with appropriate force while avoiding the application of excessive force to the human with physical limitations, is fundamental.
    Multimodal integration empowered by deep-learning technologies is a key factor to address complex force interaction, considering fabric elasticity, friction at the sock-foot interface, and the directional control of the pulling force.
    However, when it comes to the use of imitation learning, individual differences among humans inhibit accurate force interaction.
    On the other hand, using reinforcement learning has the challenge of simulating accurate interactive force.
    In this study, we formulate the problem of planning and executing the sophisticated action of putting on socks by a humanoid robot in a generalized manner that can adapt to individual differences in physique, such as human foot size, shape, color, and flexibility.

    According to the two visual streams hypothesis, the human visual system processes information through two distinct pathways: the dorsal stream for spatial awareness and action guidance, and the ventral stream for object recognition and semantic understanding \cite{visani2022semantics,van2015interactions}. 
    Recent findings emphasize the importance of interactions between these two pathways, especially in the context of complex object manipulation such as skilled grasping.
    In this study, we hypothesize that enabling a robot to assist in close-fitting garment-wearing tasks requires a combined understanding of both the object’s semantic features 
    and its spatial characteristics (e.g., depth, orientation relative to the human foot). Based on the ventral-dorsal interaction model, we posit that integrating semantic segmentation with depth estimation allows the system to plan and execute more adaptive and precise manipulations.
    To model the ventral and dorsal processing streams computationally, we utilize the Segment Anything Model 2 (SAM2) \cite{ravi2024sam2segmentimages} for semantic segmentation, which corresponds to the ventral stream’s role in object identification. For estimating depth, we adopt the Depth Anything Model (DAM) \cite{yang2024depthv2}, which mimics dorsal stream processing for spatial localization and motor planning.
    We expect that this dual-stream inspired integration will improve the robot’s ability to handle deformable, non-rigid objects in interaction with the human body, especially under conditions requiring fine motor adjustments and adaptive contact.
    
    \section{Dressing with Proposed Model}
    The overview of the proposed model is shown in Fig. \ref{fig:wide-image}.
    This model consists of multiple components and is capable of extracting temporal changes in the shape of objects such as feet and socks based on semantic understanding, without relying directly on RGB values.
    Features of the current RGB image are extracted using CNN, and the image for the next time point is reconstructed by multiplying these features with the attention map of the gaze point obtained at the previous time point using a hierarchy LSTM.
  
    \subsection{Semantic extraction of garment and human foot}
    SAM2 and DAM are applied to estimate the state of socks and the human foot. SAM2 generates semantic mask images of socks and feet, and DAM generates depth images from a monocular RGB image. 
    In general, segmentation models are vulnerable to external factors such as object deformation over time, occlusion, and varying lighting conditions \cite{ravi2024sam2segmentimages}.
    During the sock dressing, various types of occlusions frequently occur, such as the sock itself covering the foot or the robot’s arms obscuring the target area. Furthermore, we utilize the SAM2 to mitigate the influence of variations in the color and appearance of the target object (e.g., socks or feet), as well as differences in the background surrounding the object.
    The images generated by SAM2 are black and white, with brightness values of 0 and 255, where the masked regions have a brightness value of 255. In parallel, the depth images of the images acquired from the robot's camera are obtained using DAM\cite{yang2024depthv2}. The robot lacks perception in the depth direction when relying solely on two-dimensional images, so we employ DAM to infer the relative spatial relationship between the sock and the toes during insertion. Additionally, DAM facilitates estimating the vertical displacement of the arm required to successfully guide the sock over the heel region. Then, the mask depth images of the socks and feet are obtained by embedding the depth values of the depth image regions corresponding to the masked regions with a brightness value of 255.
    By adding depth information (z value) of the pixel value of the key points on the 2D coordinates of the image (x and y coordinates), 3D information on the image (x and y coordinates, and z value) is output.
    
    Let us define the following variables:
    \begin{itemize}
        \item \( D \in \mathbb{R}^{H \times W} \): The depth map obtained from the DAM.
        \item \( M \in \{0, 255\}^{H \times W} \): A binary mask image where 255 denotes the region of interest.
        \item \( \tilde{M} = \frac{M}{255} \in \{0, 1\}^{H \times W} \): The normalized mask.
        \item \( D_{\text{masked}} \in \mathbb{R}^{H \times W} \): The masked depth map where values are embedded only in the masked region.
    \end{itemize}
    We define the masked depth map as follows:
    \begin{equation}
         D_{\text{masked}}(x, y) =
        \begin{cases}
        D(x, y), & \text{if } M(x, y) = 255 \\
        0, & \text{otherwise}
        \end{cases}
    \end{equation}
    Alternatively, using the normalized mask \( \tilde{M} \), we can write:
    \begin{equation}
        D_{\text{masked}} = \tilde{M} \odot D
    \end{equation}
    where \( \odot \) denotes the Hadamard product (element-wise multiplication).
    \bigskip
    Optionally, if we prefer to mask out invalid regions using NaN instead of zero, the definition becomes:
    \begin{equation}
         D_{\text{masked}}(x, y) =
        \begin{cases}
        D(x, y), & \text{if } M(x, y) = 255 \\
        \text{NaN}, & \text{otherwise}
        \end{cases}
    \end{equation}

    \subsection{Visual and somatosensory attention mechanism} 
    The somatosensory attention mechanism facilitates the acquisition of better force interaction policy \cite{miyake2024dual}. 
    As a somatosensory attention, Selective Kernel Network (SKNet) is used to dynamically select the optimal feature extraction kernel according to the input scale using convolutions with different kernel sizes (3×3 and 5×5) \cite{li2019selective}.
    SKNet consists of three stages: Split, Fuse, and Select. In the Split-stage, the input feature map is convolved with two types of kernels. The output is integrated and pooled in the Fuse-stage. The Softmax-based weights are calculated in the Select-stage to generate the final features. This allows local and global changes of each somatosensory sensation to be captured efficiently and is effective in extracting complex changes associated with actions such as putting socks on the heels.

    As a part of visual attention mechanism, Spatial Attention Recurrent Neural Network (SARNN) is applied for robustness to background change \cite{yasutomi2023visual}. The image encoder of SARNN compresses and extracts features from vision, and the attention encoder extracts spatial attention points using the mask images as input. Multiple attention encoders enable the simultaneous estimation of attention points for multiple objects.
    To further enhance spatial understanding, we incorporate depth-aware attention. For each feature channel, a spatial softmax is applied to generate an attention map, which is used to weight the 2D spatial coordinates and depth map to calculate the expected 3D keypoints. This allows us to focus attention not only on 2D salient regions but also on the 3D geometric context, enabling robust keypoint localization even in occlusion and unknown situations.

    Let $f \in \mathbb{R}^{C \times H \times W}$ be the feature map output by a CNN, and let $d \in \mathbb{R}^{H \times W}$ be the corresponding depth map. For each channel $c \in \{1, \dots, C\}$, we compute the spatial attention map $a_c(i,j)$ by applying the softmax over the spatial dimensions:
    
    \begin{equation}
    a_c(i,j) = \frac{\exp\left( f_c(i,j) / \tau \right)}{\sum_{i',j'} \exp\left( f_c(i',j') / \tau \right)}
    \end{equation}
    where $\tau$ is a temperature parameter.
    
    The expected 2D position $(\hat{x}_c, \hat{y}_c)$ is computed as:
    
    \begin{equation}
    \hat{x}_c = \sum_{i,j} a_c(i,j) \cdot x(i,j), \quad
    \hat{y}_c = \sum_{i,j} a_c(i,j) \cdot y(i,j)
    \end{equation}
    where $x(i,j)$ and $y(i,j)$ are the spatial coordinate grids.
    
    To obtain the depth (Z-coordinate), we compute the weighted sum over the depth map using the attention map: 
    \begin{equation}
    \hat{z}_c = \sum_{i,j} a_c(i,j) \cdot d(i,j)
    \end{equation}
    Thus, the 3D keypoint for each channel $c$ is represented as:
    
    \begin{equation}
    \left( \hat{x}_c, \hat{y}_c, \hat{z}_c \right) \in \mathbb{R}^3
    \end{equation}

    \subsection{Hierarchical LSTM} 
    The six extracted 3D keypoints, joint angles, joint torques, and tactile information of the thumbs are processed by a hierarchical LSTM network to predict the next state of the sensory-motor sequence. 
    The input joint angles, joint torques, and thumb tactile information were normalized to fit into the range of 0 to 1. The maximum value and the minimum value of each parameter in the training dataset are used for normalization.
    The output of the LSTM is converted into modality information and position coordinates for the next step through a linear layer. The position information predicted by the LSTM is converted into a heat map, and the image decoder outputs the image for the next step based on the feature map and heat map. 
    Instead of processing multimodal information with a single LSTM, a more effective approach is to apply separate LSTMs to each modality. The internal representations can then be integrated through a higher-level union LSTM, allowing for better modeling of both intra-modal and inter-modal dependencies. SARNN can generalize the object's position information and learn the relationship with each modality information to generate appropriate behavior.
    At each time step $t$, our model performs two main stages: (1) updating union LSTM states using the previous hidden states from each modality, and (2) updating the bottom LSTM states for each modality using the feedback from the global context.    
    
    First, the hidden states from the previous time step for all modalities are concatenated:
    \begin{equation}
    \mathbf{u}_t^{\text{in}} = [\mathbf{h}_{t-1}^{(1)};\, \mathbf{h}_{t-1}^{(2)};\, \mathbf{h}_{t-1}^{(3)};\, \mathbf{h}_{t-1}^{(4)}] \in \mathbb{R}^{4d}
    \end{equation}
    This combined vector is passed to the union LSTM to produce the global hidden state:
    \begin{equation}
        \mathbf{h}_t^{(u)},\, \mathbf{c}_t^{(u)} = \text{LSTM}_{\text{union}}(\mathbf{u}_t^{\text{in}},\, \mathbf{h}_{t-1}^{(u)},\, \mathbf{c}_{t-1}^{(u)})
    \end{equation}
    Then, a linear layer transforms the global hidden state into a feedback vector for all modalities:
    \begin{equation}
         \mathbf{v}_t = \mathbf{W}_o \mathbf{h}_t^{(u)} + \mathbf{b}_o \in \mathbb{R}^{4d}
    \end{equation}
    The ${v}_t$ is split into modality-specific components:
    \begin{equation}
        \mathbf{v}_t = [\tilde{\mathbf{h}}_t^{(1)};\, \tilde{\mathbf{h}}_t^{(2)};\, \tilde{\mathbf{h}}_t^{(3)};\, \tilde{\mathbf{h}}_t^{(4)}]
    \end{equation}

    Second, each modality's LSTM receives its current input and the corresponding feedback hidden state from the global LSTM. It then updates its own hidden and cell states:
    \begin{equation}
        \mathbf{h}_t^{(i)},\, \mathbf{c}_t^{(i)} = \text{LSTM}_{\text{bottom}}^{(i)}(\mathbf{x}_t^{(i)},\, \tilde{\mathbf{h}}_t^{(i)},\, \mathbf{c}_{t-1}^{(i)})
        \quad i = 1, 2, 3, 4
    \end{equation}
   
    \subsection{Training and inference phase}
    In the training phase, by comparing the multimodal information predicted by the LSTM—such as joint angles, torques, tactile feedback, and attention points with the corresponding target values, the system autonomously refines its internal dynamics.
    
    The overall loss function \( L_{train} \) is expressed as follows:
    \begin{equation}
        L_{\text{train}} = \alpha L_{\text{img}} + \beta L_{angle} + \gamma L_{torque} + \delta L_{tactile} + \epsilon L_{pt}
    \end{equation}
    where \( L_{angle} \), \( L_{torque} \), \( L_{tactile} \), \( L_{img} \), and \( L_{pt} \) represent the mean squared error (MSE) of the joint angles, the joint torques, the tactile values of the thumbs, the predicted image, and the attention point \( pt = (x_t, y_t) \);
    \( \alpha \), \(\epsilon\)\, \( \beta \), \( \gamma \), and \(\delta \)\ represent the loss contributions set to 0.1, 0.1, 1.5, 1.0, and 0.2, respectively;
    Adam was used as the optimization model, and the model was trained for a total of 10,000 epochs.
    
    During motion generation, the target object is tracked based on the initial prompt information, and in addition to generating a mask image in real time, a mask depth image is generated by applying the depth image obtained from the camera image input to the DAM to the mask region. The model predicts the next state from the current joint angles, joint torques, tactile information of the thumbs, and camera images. The robot's target posture is defined based on the predicted joint angles to generate motion online.

    \section{Experiment}
    \begin{figure}[t]
     \vspace*{0.1cm}
    \centering
    \includegraphics[width=0.44\textwidth]{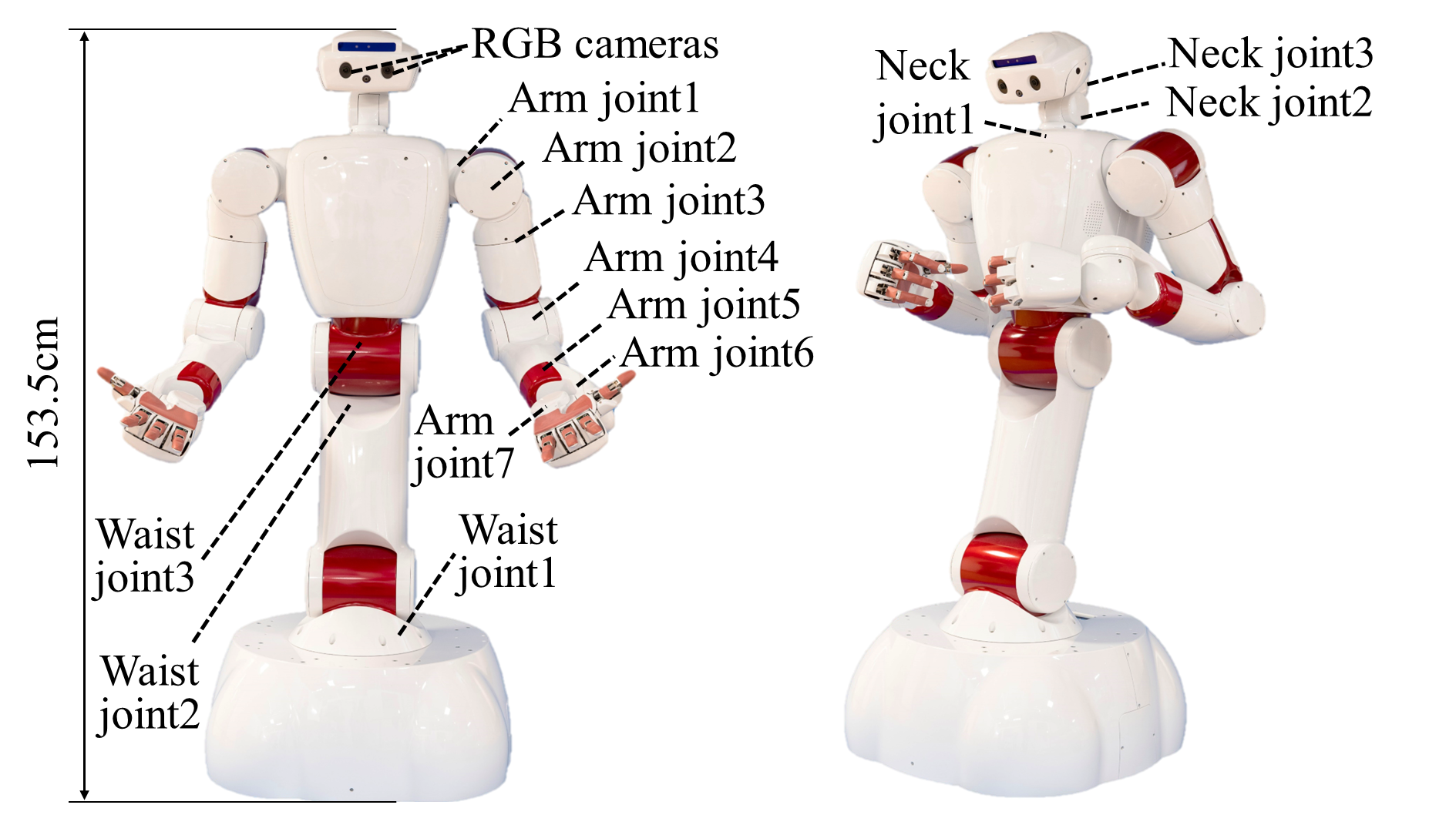} 
     \vspace*{-0.3cm}
    \caption{Robot constitutions.}
    \label{fig:airec}
     \vspace*{-0.3cm}
    \end{figure}
    
    \begin{figure}[t]
    \centering
    \includegraphics[trim=0 150 0 150, clip, width=0.4\textwidth]{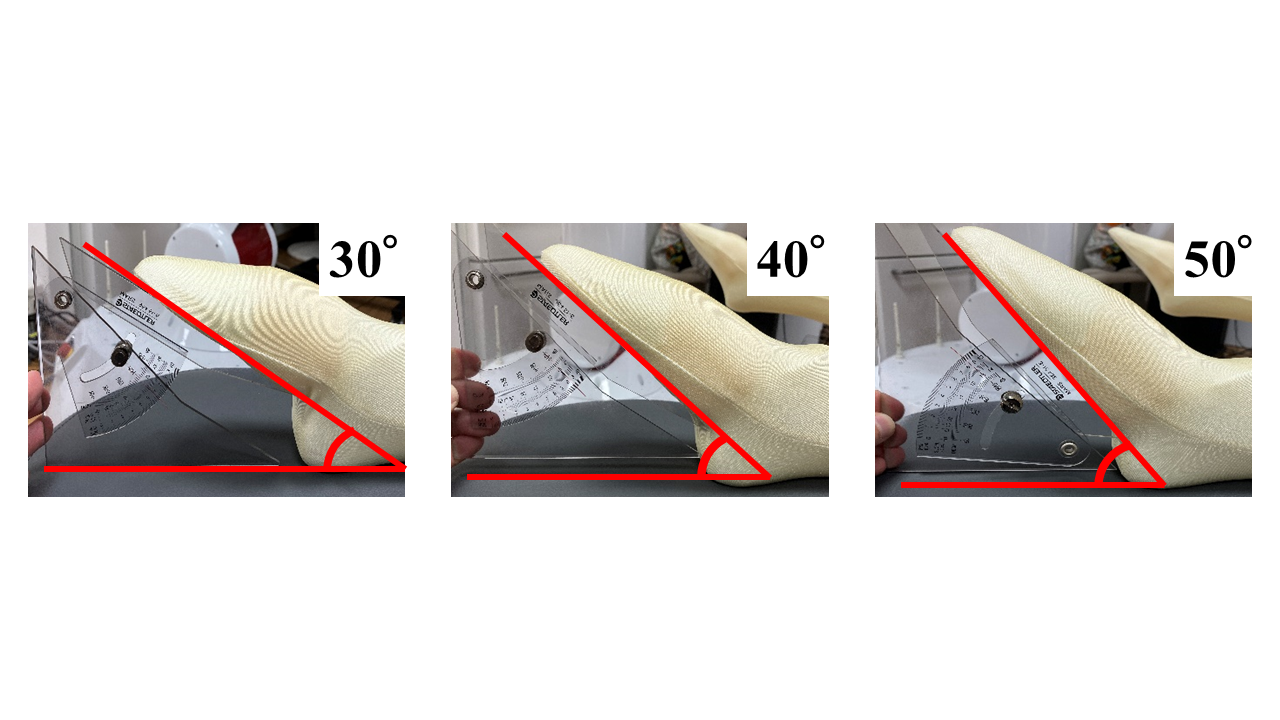} 
       \vspace*{-0.5cm}
    \caption{Condition for collecting data.}
    \label{fig:condition}
       \vspace*{-0.5cm}
    \end{figure}
  
    \subsection{Robot constitutions}
    In this study, we used a humanoid robot AIREC (Fig. \ref{fig:airec}) \cite{miyake2022skeleton}.
    It integrates various sensing functions such as torque sensors, tactile sensors, and visual sensors, making it a platform for studying robot interaction with the real world.
    In addition, the control system is highly versatile, supporting various modes such as position, force, and impedance control, and parameters such as mass, damping, and spring can be customized. This adaptability allows the robot to perform tasks while improving safety, enhancing safety during contact with humans. Each arm of the robot has seven joints, and impedance controllers are used for the joints in the hand that may come into contact with the robot to soften them and improve safety. 
    The robot incorporates hardware that facilitates reaching lower positions via waist bending.

    \subsection{Demonstration with PS controller}
    The robot was teleoperated using a PS controller.
    Hand positions were updated by adding xyz displacements from the controller inputs, and target joint angles were computed via inverse kinematics.
    Two motion patterns were used: parallel arm motion with identical displacements, and alternating arm motion in opposite directions.
    The parallel pattern, being intuitive and easy to reproduce, was primarily used for generating consistent demonstration data, while the alternating pattern was introduced to handle more diverse motions caused by friction or snagging.

    \subsection{Environment and dataset}
\begin{figure}
        \centering
        \vspace*{0.2cm}\includegraphics[trim={0 0.8cm 0 1.5cm},clip,width=0.74\linewidth]{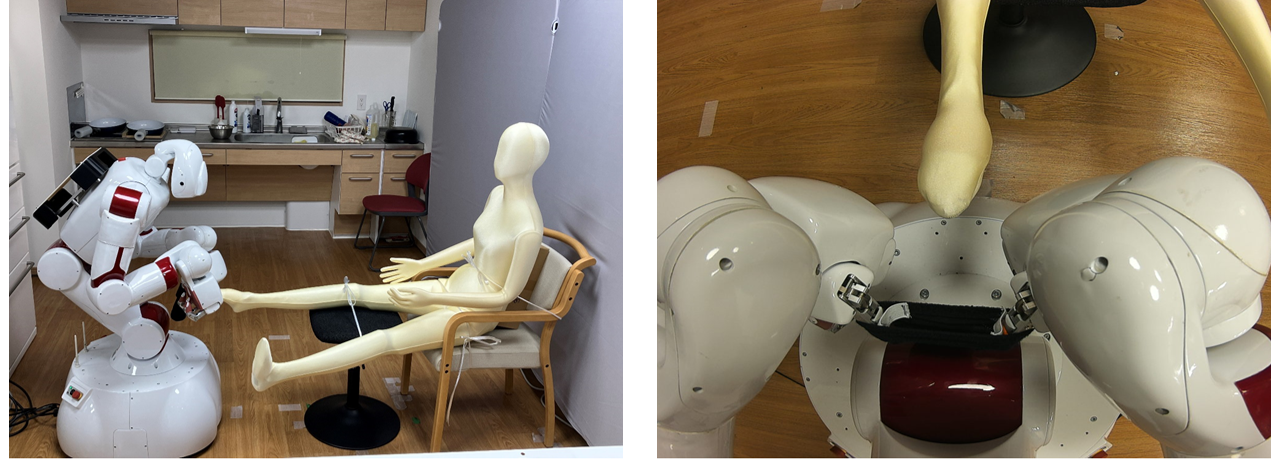}
        \vspace*{-0.27cm}
        \caption{Experimental setup (left) and robot’s view (right).}
        \label{fig:train_env}
       \vspace*{-0.63cm}
    \end{figure}

    In this study, a mannequin was used as a training subject. The specifications of the mannequin are 170 cm in height, 40 cm in shoulder width, and 19 cm in foot size. The exterior is made of urethane, and the interior is made of polystyrene foam.
    The socks were composed of rayon (66\%), polyester (29\%), silk (3\%), and polyurethane (2\%), with a size of 24–26 cm and a thickness of 0.9 mm. Stiffness was lower longitudinally (stress = 0.094 MPa, Young’s modulus = 0.103 MPa at 90.9\% elongation) and higher transversely (stress = 0.497 MPa, Young’s modulus = 0.497 MPa at 100\% elongation). 
    As shown in Fig.~\ref{fig:train_env}, the mannequin was seated in a chair and foot was on a footrest. The seat height of the chair and the footrest were set to 45 cm. We assume a scenario in which a robot dresses socks for a person whose legs are out of bed and whose feet are off the floor. The participants maintained their natural sitting posture, and no quantitative assessment of posture was conducted.

    The mannequin is lightweight, allowing its leg joints to bend easily. The mannequin's feet are fixed with strings during training to apply the movements to real humans. 
    Putting on socks requires balance,lower-body strength, and flexibility, which makes it a difficult movement for people with physical limitations.
    Collecting datasets in advance also places a burden on human participants.
    Therefore, a mannequin was used in the training phase for safety, and the adaptability of the model to humans was evaluated. 
  
    As shown in Fig.~\ref{fig:condition}, since the feet position varies from person to person, we set the mannequin’s foot angle at three different positions—30°, 40°, and 50° from the horizontal plane.
    To balance the computation cost and training efficiency, we collected 4 samples for each, resulting in 12 samples for training.
    The algorithm was set to a frequency of 20 Hz, resulting in 210 time steps per dataset.
    The dataset includes the robot's vision, the joint angles and torques of both arms' joints 1-7, and the tactile information of the thumb, as shown in Fig.~\ref{fig:airec}. The initial joint angles of the head were set to 0°, 37°, and 0° for head/joint1, head/joint2, and head/joint3, respectively. Similarly, the initial angles for the torso were set to 0°, -45°, and 100° for torso/joint1, torso/joint2, and torso/joint3, respectively.
    An impedance controller is applied to joint 7 to ensure safety in scenarios where contact with the human body may occur. The other joints employ position control, as actuation force is required to manipulate the sock effectively during the dressing process. In this study, the tactile information of the thumb is used to grasp a sock with the thumb and recognize the tightness of the sock.
    
    \subsection{Evaluation}
    We conducted two types of evaluation experiments to evaluate the overall system performance of the proposed model.
    First, we conducted an ablation study to evaluate the contribution of each component to the model’s performance. 
    Second, we investigated whether the model trained using mannequin data can generalize to variations in individual subjects and environmental conditions. We compared its performance with existing models to evaluate its robustness.
    \subsubsection{Ablation Study of Model Architecture Components}
    We conducted four experiments selectively removing key components: hierarchical LSTM, SKNet, DAM, and both SAM and DAM. We evaluated the effectiveness and contribution of semantic-based visual attention mechanism.
    
    As an evaluation criterion common to both experiments, 300 loops were performed during inference, and the system was regarded successful if the following was achieved at the end of the loop.
    It was regarded successful if the sock was inserted in the toe of the mannequin without the arms interfering with each other, passed through the heel, reached the ankle, and maintained that state. 
    
    \subsubsection{Evaluation of generalization capability and robustness}
    To evaluate the generalization performance and robustness of the proposed model, we conducted experiments on 10 real humans and compared it with two existing models:
 
    • Action Chunking with Transformer (ACT) \cite{zhao2023learning} is a method that divides a long-term action sequence into coherent chunks and models them using Transformer. Since ACT utilizes high-level context information, it is possible to plan actions that take into account long-term dependencies.
   
    • Diffusion Policy (DP) \cite{chi2024diffusionpolicyvisuomotorpolicy} is a method of generating robot behavior using a diffusion model, and is excellent at generating continuous movements that combine diversity and smoothness. Since it can generate multiple behavior trajectories through probabilistic prediction, it is effective for tasks that require a high degree of exploration. 
    
    The study involved ten participants with foot sizes between 23.0 cm and 26.5 cm. The model’s adaptability to environmental changes and individual differences in foot characteristics was assessed. Each method was evaluated in 50 trials per condition. Ethical approval was obtained from the Human Subject Research Ethics Review Committee of Waseda University (No. 2022-485).

    \section{Results and discussion}
    \subsection{Ablation study on model components}

    In Table~\ref{tab:ablation}, Dress-success and Insert-success denote the success rate of the complete dressing after toe-insertion and toe-insertion success rate, respectively.  
    Removal of DAM or SKNet caused moderate performance degradation, while removal of Hierarchical LSTM or SAM2 and DAM caused a dramatic drop in dress-success rate, respectively. 
    Without DAM, the model achieved a 85\% success rate. However, one failure case was observed: after toe-insertion, when the arm was lowered along the surface of the feet, the toe got caught in the sock, making it difficult to continue movement. 
    
    Moreover, experiments validating the toe-insertion process revealed that the proposed model achieved the highest success rate compared with its ablated variants, as shown in Table \ref{tab:ablation}, indicating superior robustness in handling the insertion phase. However, four failures due to slight misalignments suggest that the system remains sensitive to positional differences.
    The insertion time was approximately 4 s.
    These results confirm the critical role of temporal modeling and the importance of spatial and depth-aware attention mechanisms.

    \begin{table}[t]
      \centering
      \vspace*{0.2cm}
      \caption{Ablation study of model architecture}
      \vspace*{-0.2cm}
      \label{tab:ablation}
      \begin{tabular}{lcc}
        \toprule
        \textbf{Model Variant} & \textbf{Dress-success rate} &  \textbf{Insert-success rate} \\
        \midrule
        Ours                     & \textbf{20/20} & \textbf{16/20}\\
        No DAM                   & 17/20 & 6/20\\
        No SKNet                 & 16/20 & 8/20\\
        No Hierarchical LSTM     & 1/20  & 2/20\\
        No SAM and DAM           & 0/20  & 0/20\\
        \bottomrule
      \end{tabular}
    \end{table}
    
    \subsection{Results of generalization capability and robustness}
    
    \begin{figure}[h]
     \centering
\includegraphics[width=0.38\textwidth]{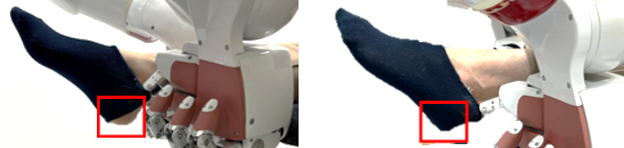}
    \caption{Example failure (left) and success (right) scenes.}
    \label{fig:motion-known-unknown}
   \vspace*{-0.2cm}
    \end{figure}

    \begin{table}[t]
      \centering
      \caption{Success rates of motion generation.}
   \vspace*{-0.3cm}
      \label{tab:generalization}
      \begin{tabular}{lccc}
        \toprule
        \textbf{Condition} & \textbf{ACT} & \textbf{DP} & \textbf{Ours} \\
        \midrule
        Known Background   & 33/50        & N/A         & \textbf{42/50} \\
        Unknown Background & 0/50         & N/A         & \textbf{37/50} \\
        \bottomrule
      \end{tabular}
         \vspace*{-0.48cm}
    \end{table}
    
    \begin{figure}
        \centering
         \vspace*{-0.3cm}
     \includegraphics[width=\linewidth]{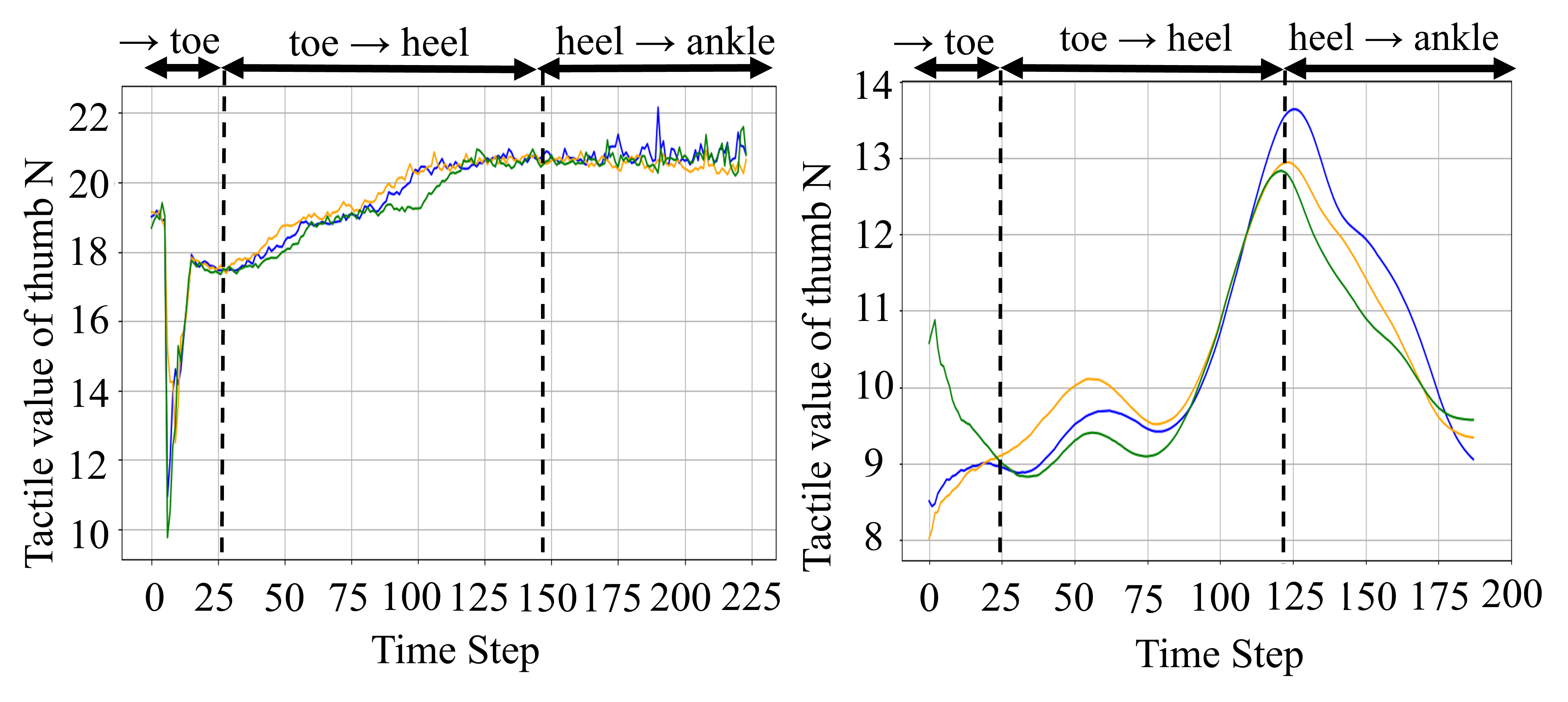}
   \vspace*{-0.6cm}
        \caption{Example of time series changes in thumb tactile values. The blue, orange, and green lines indicate foot sizes of 26.0, 25.0, and 24.0, respectively, for the proposed method (left) and the ACT (right).}
        \label{fig:tactile}
           \vspace*{-0.1cm}
    \end{figure}
    
    \begin{table}[t]
      \centering
      \vspace*{0.2cm}
      \caption{Maximum Tactile Forces for different foot sizes and phases}
       \vspace*{-0.2cm}
      \label{tab:tactile_forces}
      \begin{tabular}{lcccccc}
        \toprule
        \multirow{2}{*}{\textbf{Foot Size [cm]}} 
          & \multicolumn{2}{c}{\textbf{toe [N]}} 
          & \multicolumn{2}{c}{\textbf{toe--heel [N]}} 
          & \multicolumn{2}{c}{\textbf{heel--ankle [N]}} \\
        \cmidrule(lr){2-3} \cmidrule(lr){4-5} \cmidrule(lr){6-7}
          & Ours & ACT & Ours & ACT & Ours & ACT \\
        \midrule
        24.0 & 19.08 & 10.88 & 20.94 & 12.83 & 21.83 & 12.59 \\
        25.0 & 19.20 & 9.11 & 20.87 & 12.94 & 20.93 & 12.88 \\
        26.0 & 19.19 & 9.01 & 20.96 & 13.63 & 22.16 & 13.63 \\
        \bottomrule
      \end{tabular}
       \vspace*{-0.1cm}
    \end{table}

    Table~\ref{tab:generalization} shows that our method achieved a success rate of \textbf{84\%} (42/50) in the trained background and \textbf{74\%} (37/50) in the untrained background. In contrast, ACT achieved 66\% (33/50) and 0\% (0/50), respectively, and DP failed to complete the dressing task (Fig.~\ref{fig:motion-known-unknown}). Our method could mitigate environmental interference, such as lighting and shadow variations, although such factors still affect performance.
    
    Fig.~\ref{fig:tactile} shows the time series of thumb tactile values for different foot sizes. 
    In Table~\ref{tab:tactile_forces}, the proposed model demonstrated a gradual increase in tactile force throughout the phases: \(\approx19\)~N at the toe, \(\approx21\)~N at toe--heel, and \(\approx22\)~N at heel--ankle, 
    exhibiting stable outputs across foot sizes. 
    In contrast, the ACT model showed a different pattern, with minimal force at the toe phase and 
    a maximum at the toe--heel. 
    The ACT was more sensitive to size variations than the proposed method. 
    The larger force at the heel--ankle phase in the proposed model was likely due to the deeper arm position and the wider hand spacing required to reach the heel area, enhancing tactile interaction. 
    The consistent force across different foot sizes suggested that the proposed model effectively maintained an appropriate level of contact force to prevent excessive pressure during the dressing motion.
    
    Human experiments revealed failure modes that were not observed in mannequin tests, primarily arising from morphological variability. Limited ankle flexibility and toenail snagging caused difficulties during toe insertion. 
    Moreover, as shown in Table~\ref{tab:foot_size}, foot size caused specific failures. The participants with larger feet experienced heel-catching failures due to premature arm lifting (Fig.~\ref{fig:motion-known-unknown}).
    Consequently, although the system can accommodate a certain range of foot sizes, challenges remain in achieving robust performance across broader morphological variations.
    We would expand the method by leveraging a morpho-aware method \cite{wang2021reconfigurability}.

\begin{figure}[h]
        \centering
         \vspace*{0.2cm}       
        \includegraphics[width=0.4\textwidth]{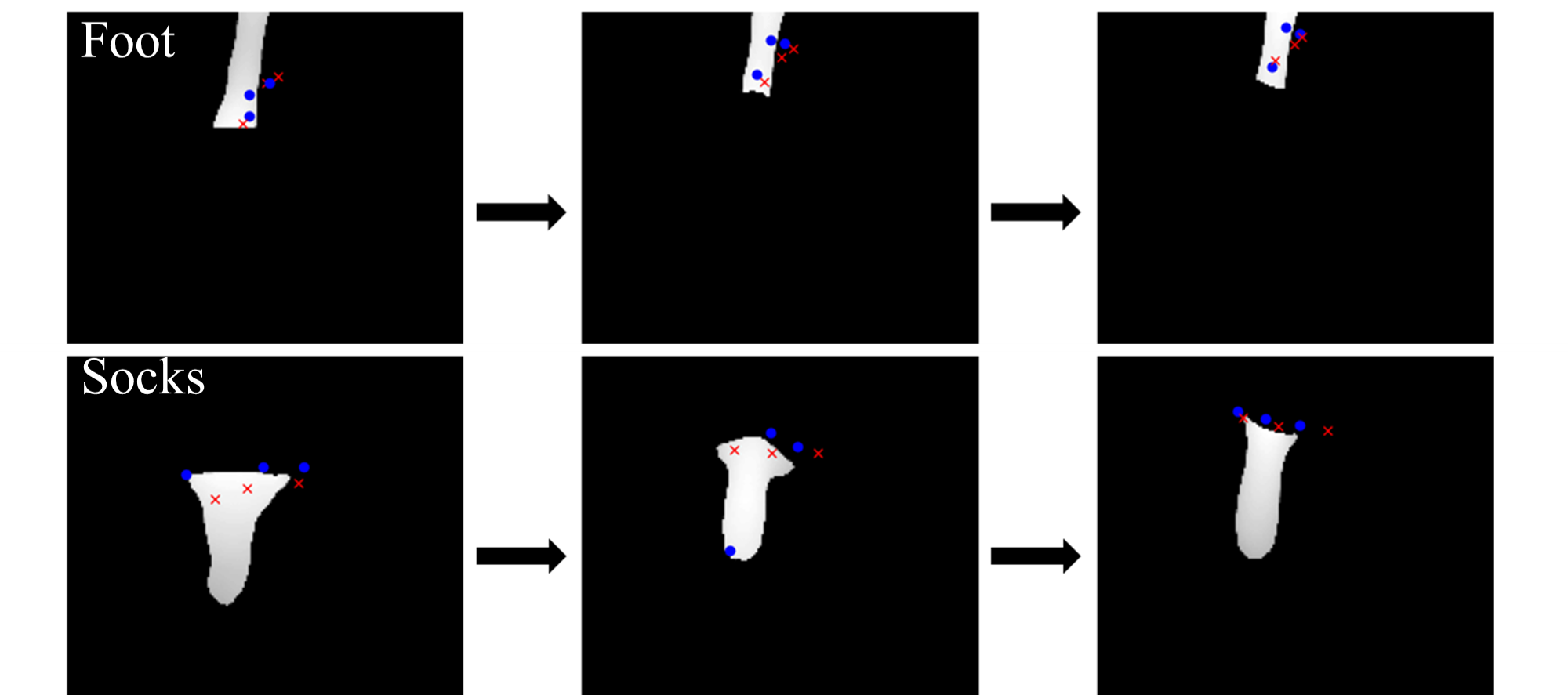}
        \vspace*{-0.24cm}        
        \caption{Attention points moving over the object area. Blue points and red points indicate current and future image attention key points, respectively.}
        \vspace*{-0.63cm}       
        \label{fig:attentionpoint_object}
    \end{figure}

\begin{figure*}
         \centering
   \vspace*{0.1cm}    \includegraphics[width=0.812\textwidth]{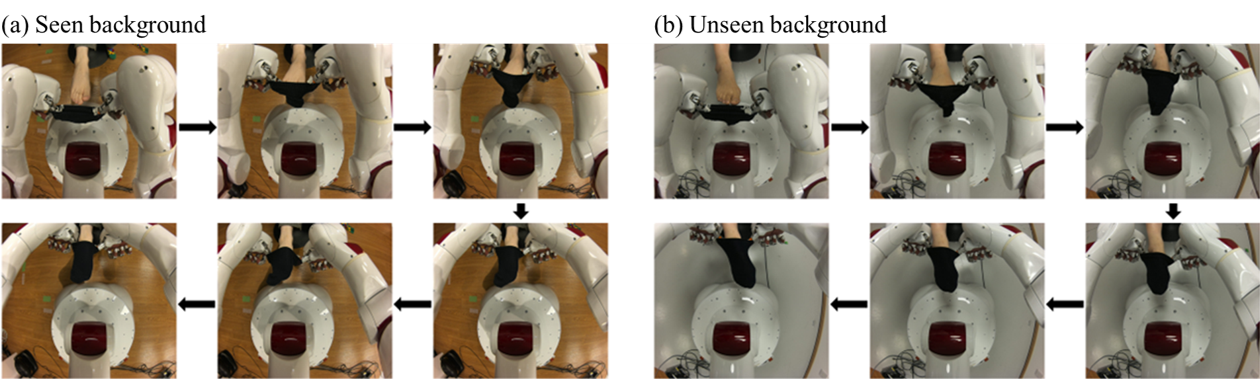}
   \vspace*{-0.3cm}
    \caption{Scenes of the test of the motion generation with the proposed model; seen background(left) unseen background(right)}
   \vspace*{-0.6cm}
    \label{fig:image_by_50steps}
    \end{figure*}

\begin{figure}[h]
             \centering            \includegraphics[width=0.42\textwidth]{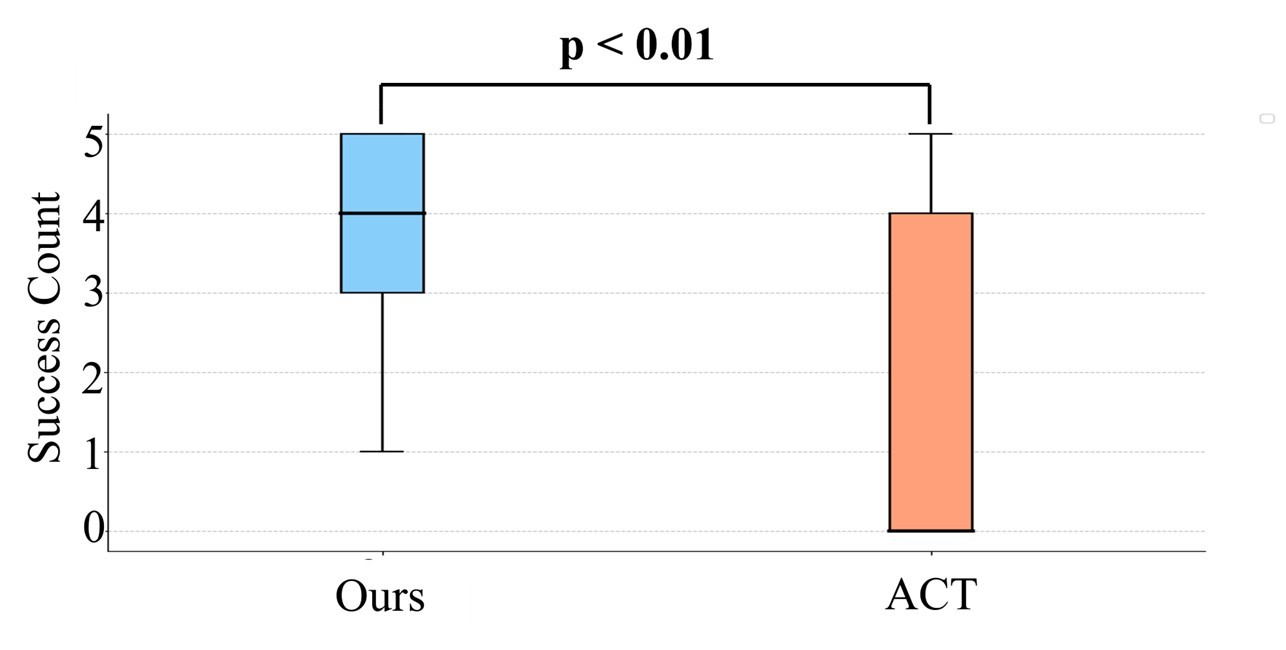}
       \vspace*{-0.35cm}
            \caption{Boxplot of number of successes. Our method demonstrated significantly higher performance than ACT.}
            \label{fig:Boxplot of number of successes}
       \vspace*{-0.65cm}
    \end{figure}

    A scene of attention points moving over the object area of semantic mask images is shown in Fig.~\ref{fig:attentionpoint_object}. The attention points consistently track the object area.
    The proposed model was robust to unseen environments and individual anatomical variations, as shown in Fig. ~\ref{fig:image_by_50steps}. 
    The integration of semantic and somatosensory processing was effective, achieving efficient learning even with limited data.
    If the somatosensory modality were absent, the model would not function even with larger datasets. 
    Fig.~\ref{fig:Boxplot of number of successes} shows the comparison results of the online performance between our model and ACT.
    This box plot shows the distribution of the number of successes out of 5 for 10 participants. We conducted a Wilcoxon signed rank test (Shapiro-Wilk test confirmed nonparametric) on the combined number of successes in the seen and unseen backgrounds. 
     There was a statistically significant difference and large effect size between the proposed model and ACT (p-value $<$ 0.01, Cohen’s dz = 1.11, 95\% CI = [0.64, 1.93]). 
     Although all models can perform well offline, ACT and DP trained on 12 datasets showed poor online performance due to inadequate force control, as shown in Table~\ref{tab:foot_size}.

    \begin{table}[t]
     \vspace*{0.2cm}
      \centering
      \caption{Success rates corresponding to foot sizes.}
      \label{tab:foot_size}
         \vspace*{-0.2cm}
      \begin{tabular}{lccc}
        \toprule
        \textbf{Size of foot} & \textbf{ACT} & \textbf{DP} & \textbf{Ours} \\
        \midrule
        23-24 cm  & 14/30        & N/A         & \textbf{26/30} \\
        24-25 cm  & 14/30        & N/A         & \textbf{28/30} \\
        25-26 cm  & 5/30        & N/A         & \textbf{21/30} \\
        26-27 cm  & 0/10         & N/A         & \textbf{4/10} \\
        \bottomrule
      \end{tabular}
   \vspace*{-0.2cm}
    \end{table}

Stability and reproducibility are particularly important in the task of putting on socks. In DP, the consistency of the behavior was sometimes lost due to the probabilistic nature of behavior generation, making the movement of guiding the sock to the toe difficult and oscillatory.
ACT tended to vary the arm trajectory. 
In an unknown background, ACT collided with the foot even in the first phase of reaching the foot.
We assume that this is because the self-attention mechanism in Transformer has low sensitivity to local changes, which limits adaptation to differences due to foot size differences. 
    
During the user study, socks sometimes caught on nails, making it difficult to continue the dressing motion. This necessitated motion replanning, highlighting the inherent complexity and unpredictability involved in sock-dressing assistance. Moreover, while our model performed better with limited training data, achieving a better trade-off between better generality and training cost remains future work.

    \section*{Conclusion and future work}

    A novel multimodal imitation learning method, utilizing a hierarchical LSTM, for a humanoid robot assisting with close-fitting garment dressing was developed. 
    Visual information is embedded with depth information in a semantic mask image, and the system recognizes the changing shapes of the foot and socks. 
    The model trained on a mannequin's foot demonstrated higher performance than the baseline in a subject experiment with 10 participants, taking into account various foot sizes, foot widths, and flexibility. 
    
    Although our system is capable of performing the insertion motion, the precision of that phase remains limited, and a rigorous evaluation of the toe-insertion step is outside the scope of this study. Instead, we concentrate on learning and executing robust movements in the subsequent phase, where the robot must adapt to differences in foot geometry and sock deformation.
    In future work, we would extend the system to handle dynamic adaptation when unexpected foot movements occur during the dressing process. In particular, the ability to replan motions online based on the detection of unintended contact or misalignment could further enhance the robustness and flexibility of the system in real-world applications.
 
    \bibliographystyle{ieeetr}
    \bibliography{root}
\end{document}